\pdfoutput=1
\documentclass[11pt]{article}

\usepackage[final]{acl}

\usepackage{times}
\usepackage{latexsym}
\usepackage{multirow}
\usepackage{booktabs}
\usepackage{subcaption}
\usepackage{graphicx}
\usepackage{cleveref}
\usepackage{comment}
\usepackage[T1]{fontenc}
\usepackage[utf8]{inputenc}

\usepackage{microtype}

\usepackage{inconsolata}
\usepackage{xspace}
\newcommand*\samethanks[1][\value{footnote}]{\footnotemark[#1]}
\newcommand*{\ekohate}{\textsc{EkoHate}\xspace}

\title{EkoHate: Abusive Language and Hate Speech Detection for Code-switched Political Discussions on Nigerian Twitter}

\author{\normalsize Comfort Eseohen Ilevbare\textsuperscript{1}\thanks{Equal contribution.}, \ Jesujoba Oluwadara Alabi\textsuperscript{2}\samethanks, \  David Ifeoluwa Adelani\textsuperscript{3}, \\ 
\textbf{\normalsize Firdous Damilola Bakare\textsuperscript{1}, Oluwatoyin Bunmi Abiola\textsuperscript{1} and Oluwaseyi Adesina Adeyemo\textsuperscript{1}} \\
\textsuperscript{1} Department of Computer Science, Afe Babalola University, Ado-Ekiti, Nigeria  \\
\textsuperscript{2} Spoken Language Systems, Saarland University,
Saarland Informatics Campus, Germany\\
\textsuperscript{3} University College London \\
\texttt{jalabi@lsv.uni-saarland.de}, \texttt{d.adelani@ucl.ac.uk}\\
\texttt{\{abiolaob,adeyemo\}@abuad.edu.ng} 
}

\begin{document}
\maketitle
\begin{abstract}
Nigerians have a notable online presence and actively discuss political and topical matters. This was particularly evident throughout the 2023 general election, where Twitter was used for campaigning, fact-checking and verification, and even positive and negative discourse. However, little or none has been done in the detection of abusive language and hate speech in Nigeria. In this paper, we curated \textit{code-switched} Twitter data directed at three musketeers of the governorship election on the most populous and economically vibrant state in Nigeria; Lagos state, with the view to detect offensive speech in political discussions. We developed \ekohate---an abusive language and hate speech dataset for political discussions between the three candidates and their followers using a binary  (normal vs offensive) and fine-grained four-label annotation scheme. We analysed our dataset and provided an empirical evaluation of state-of-the-art methods across both supervised and cross-lingual transfer learning settings. In the supervised setting, our evaluation results in both binary and four-label annotation schemes show that we can achieve 95.1 and 70.3 F1 points respectively. Furthermore, we show that our dataset adequately transfers very well to three publicly available offensive datasets (OLID, HateUS2020, and FountaHate), generalizing to political discussions in other regions like the US.  %with at least 62.7 F1 points. 

%code-switched data was sourced from tweets directed at three musketeers of the governorship election of the most populous and economically vibrant state in Nigeria; Lagos state, with the view to detect offensive and hate speech on political discussion. Our annotation of the language used in our data shows that $63.1\%$ of the data is from English, while the others are either in Yoruba ($3.5\%$), Nigerian-Pidgin ($7.3\%$), or Code-switched ($26.0\%$) between English and the other two languages (Nigerian-Pidgin and Yoruba).
\end{abstract}

\section{Introduction}
The internet, with various social media platforms, has interconnected our world, facilitating real-time communication. One area that has benefited from the use of social media platforms is elections at various levels. Research has shown that these platforms have an impact on the outcome of elections in different countries~\citep{NBERw28849,carney2022effect}, but not without the spread of false information~\citep{Grinberg_2019,matt_carlson_2020,turgay_2020}, dissemination of hate speech~\citep{QJPS_19045,nwozor2022social}, and various other forms of attacks. Therefore, efforts have been made to automatically identify hateful and divisive comments~\cite {hateoffensive}. They include supervised methods, that focus on  %which involve  
curating hate speech datasets~\citep{Mathew_Saha_Yimam_2021, demus-etal-2022-comprehensive, piot2024metahate}.

However, the majority of these datasets were created for elections in the US~\citep{suryawanshi-etal-2020-multimodal, grimminger-klinger-2021-hate, Zahrah_2022} and other non-African countries~\citep{Alfina_2017, Febriana_2019}. In this work, we focus on Nigerian elections. Nigerians have a notable online presence and actively discuss political and topical matters. This was particularly evident throughout the 2023 general election, where Twitter was used for campaigning, fact-checking, verification, and positive and negative discourse. However, little or none has been done in the detection of offensive and hate speech in Nigeria. % as individuals frequently utilize local languages, tones and forms of communication, such as Pidgin English, Yoruba and other dialects on social media~\citep{hate}. This emphasizes the need to perform the analysis on a locally sourced dataset for validation and model applicability to the local environment.

\begin{table*}[t]
 \begin{center}
 \footnotesize
 \scalebox{0.90}{
 \begin{tabular}{l|ccc}
 \toprule
 \textbf{Tweet} &  \textbf{N-O} & \textbf{N-A-H-C} \\
\midrule
Bro, go to the field and gather momentum. Social media can only do so much & N & N \\ 
LOL. This guy na mumu honestly & O & A \\
%Oga gettout, you go no where, you and ur principal r just mere opportunist & O & A \\ 
% You wey not even know your left from your right. You sound like you're running for the post of SUG president. & O & A \\ 
A bl00dy immigrant calling another person immigrant... & O & H \\
You will still be voted out of office sir.  & - & C \\
\bottomrule
\end{tabular}
}
  \vspace{-2mm}
  \caption{Examples of tweets and their labels under two labelling schemes. In the second example ``na mumu'' can mean ``is a fool" . N is Normal, O is offensive (i.e. Abusive \& Hateful), A is abusive, and C is contempt.   \looseness-1}
  \label{tab:examples}
  \end{center}
\end{table*}

In this paper, we create \ekohate---a new code-switched abusive language and hate speech detection dataset containing 3,398 annotated tweets gathered from the posts and replies of three leading political candidates in Lagos, annotated using a binary (``normal'' vs ``offensive'' i.e abusive \& hateful) and fine-grained four-label annotation scheme. The four-label annotation scheme categorizes tweets into ``normal'', ``abusive'', ``hateful'', and ``contempt''. The last category was added based on the difficulty to classify some tweets that do not properly fit into ``normal'' or ``abusive'' but express strong disliking in a neural tone, suggested by \citep{ron-etal-2023-factoring}.  \autoref{tab:examples} shows some examples of tweets and their categorization. The last example ``You will still be voted out of office sir.'' does not fit the categorization of ``offensive'' but can be ``contemptuous'' to a sitting Governor, implying that despite his campaign, he would still be voted out.% It is mindful to note that additional annotation was conducted to label the language of each tweet, distinguishing amongst English, Yoruba, Nigerian Pidgin, or code-switching instances.

Our evaluation shows that we can identify the offensive tweets with the high performance of $95.1$ F1 by fine-tuning a domain-specific Twitter BERT model~\cite{barbieri-etal-2020-tweeteval}. However, on a four-label annotation scheme, the F1-score drops to $70.3$ F1 showing the difficulty of the fine-grained labeling scheme. Furthermore, we conduct cross-corpus transfer learning experiments using OLID~\citep{zampieri-etal-2019-predicting},  HateUS2020~\citep{grimminger-klinger-2021-hate}, and FountaHate~\citep{Founta_Djouvas_paper} which achieved $71.1$ F1, $58.6$ F1, and $43.9$ F1 points respectively on \ekohate test set. Interestingly, we find that our dataset achieves a good transfer performance to the existing datasets reaching an F1-score of $71.8$ on OLID, $62.7$ F1 on HateUS2020 and $53.6$ on FountaHate, which shows that our annotated dataset generalizes to political discussions in other regions like the US despite the cultural specificity and code-switched nature of our dataset. We hope our dataset encourages the evaluation of hate speech detection methods in diverse countries. The data and code are available on GitHub\footnote{\url{https://github.com/befittingcrown/EkoHate}}

%We annotated about 3,398 tweets gathered from the posts and replies of three leading political candidates in Lagos.

%code-switched data was sourced from tweets directed at three musketeers of the governorship election of the most populous and economically vibrant state in Nigeria; Lagos state, with the view to detect offensive and hate speech on political discussion. Our annotation of the language used in our data shows that $63.1\%$ of the data is from English, while the others are either in Yoruba ($3.5\%$), Nigerian-Pidgin ($7.3\%$), or Code-switched ($26.0\%$) between English and the other two languages (Nigerian-Pidgin and Yoruba).

%What proportion of each language is abusive, hateful, or contemptuous?

\section{\ekohate dataset}
\subsection{Lagos Gubernatorial Elections}

Lagos (also known as \`{E}k\'{o}) is the commercial nerve centre of Nigeria, the former federal capital of Nigeria, and the most populous city in Nigeria and Africa with over 15 million residents according to \citet{doris_doris_sasu_stat}. In the 2023 Nigerian election, Lagos is probably the most strategic state because of its voting power, and most importantly because the leading candidate for the presidential election is from Lagos. There were three leading candidates from the major political parties: All Progressives Congress (APC), Peoples Democratic Party (PDP), and Labour Party (LP). The latter was particularly popular on social media and especially among the youths because Nigerians saw it as a third force. Therefore, there was a lot of controversial and offensive tweets on social media during the election of Lagos. Thus, we focus on analyzing the political tweets during the last Lagos election. 

\begin{figure}[t]
\setlength{\belowcaptionskip}{-10pt}
    \centering   \includegraphics[width=0.45\textwidth]{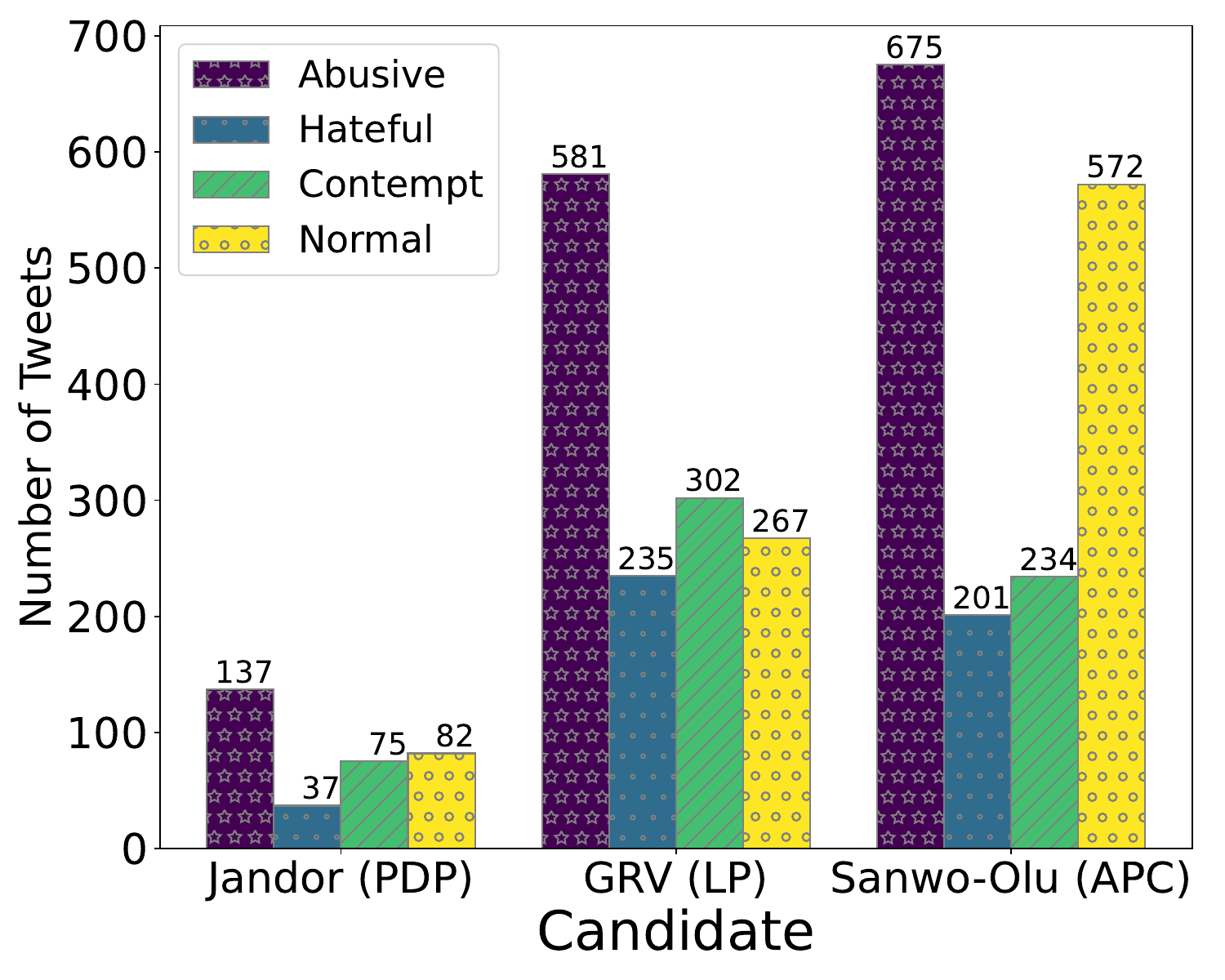}
 \vspace{-2mm}
    \caption{\textbf{EkoHate}: The distribution of the classes per candidate. \looseness-1}
    \label{fig:candidate_chart}
\end{figure}

% \begin{table}[t]
% \centering
% \footnotesize
% \begin{tabular}{c|r|rrrr|r}
%    \toprule
%    & \textbf{No. } & \multicolumn{5}{c}{\textbf{Label}} \\
%    \textbf{Candidate} & \textbf{Tweets} & \textbf{N} & \textbf{C} & \textbf{A} & \textbf{H} & \textbf{O} \\
%    \midrule
%    Sanwo-Olu & 1682 & 572 & 234 &675  &201  & 876   \\
%    GRV & 1385 & 267 & 302 & 581 & 235 &   816  \\
%    Jandor & 331 & 82  & 75  & 137  & 37 & 174 \\
%    \bottomrule
%  \end{tabular}
%  \vspace{-1mm}
%  \caption{\textbf{EkoHate}: Data source, number of movie reviews per source, and average length of reviews }
%  \label{tab:data_stat}
%\end{table}

\subsection{Labelling Scheme}
There are different labeling scheme for offensive and hate-speech on social media. The simplest approach is to categorize the tweets as either \textit{offensive} or \textit{non-offensive}~\citep{zampieri-etal-2019-predicting}. In the literature~\citep{hateoffensive,Founta_Djouvas_paper}, it is popular to distinguish between the type of \textbf{offensive} content as either \textit{abusive} or \textit{hateful}. Here, we adopted the labelling scheme of \textbf{normal} (or non-offensive), \textbf{abusive}, \textbf{hateful}, and \textbf{contempt}. The last one was added based on the difficultly of accurately classifying some political tweets showing a strong disliking to someone but expressed using a neutral tone, following the categorization of \citet{ron-etal-2023-factoring}. Examples of such tweets are: ``Just dey play oooo'' and ``The sheer effrontery! (..to be contesting)'', ``As if we were sitting before'' (a response to---\`{E}k\'{o} E dìde (stand up Lagos)!!  ~ GRV..).

\paragraph{Anotators} The annotators consist of two female individuals: one undergraduate and one postgraduate student in computer science. Neither annotator is from Lagos state nor affiliated with any of the political parties. They underwent a training session for the task, which involved introducing them to the task and %familiarizing them with the 
Label Studio\footnote{\url{https://labelstud.io/}} annotation platform.

\paragraph{Data collection and Annotation}

Tweets were manually extracted from twitter platform over a  period of ten weeks and about 3,398 tweets were collected and annotated.  For the purpose of this study, only tweets and replies from three candidates—Babajide Olusola Sanwo-Olu representing APC, Gbadebo Chinedu Patrick Rhodes-Vivour popularly known as GRV representing LP, and Abdul-Azeez Olajide Adediran, popularly known as Jandor representing PDP, were utilized due to the substantial traffic and reactions on their pages, providing ample data for this research. The corpus was annotated by two volunteers for the following five different label categories, \textit{normal},  \textit{contempt}, \textit{abusive}, and \textit{hateful} and \textit{indeterminate}. None of the tweets were classified as indeterminate. The inter-agreement score of the annotation in terms of \textbf{Fleiss Kappa} score is \textbf{0.43} signifying a moderate agreement. Since, we only have two annotators, we could not use majority voting. To determine the final annotation, we ask the two to meet in-person, discuss and resolve the conflicting annotations.   %Due to the difficultAfter annotation, %the two annotators met to discuss and resolve conflicting annotation. 
Finally, one of the authors of the paper did a review of the annotation to check for consistency.

\begin{figure}[t]
\setlength{\belowcaptionskip}{-10pt}
    \centering   \includegraphics[width=0.40\textwidth]{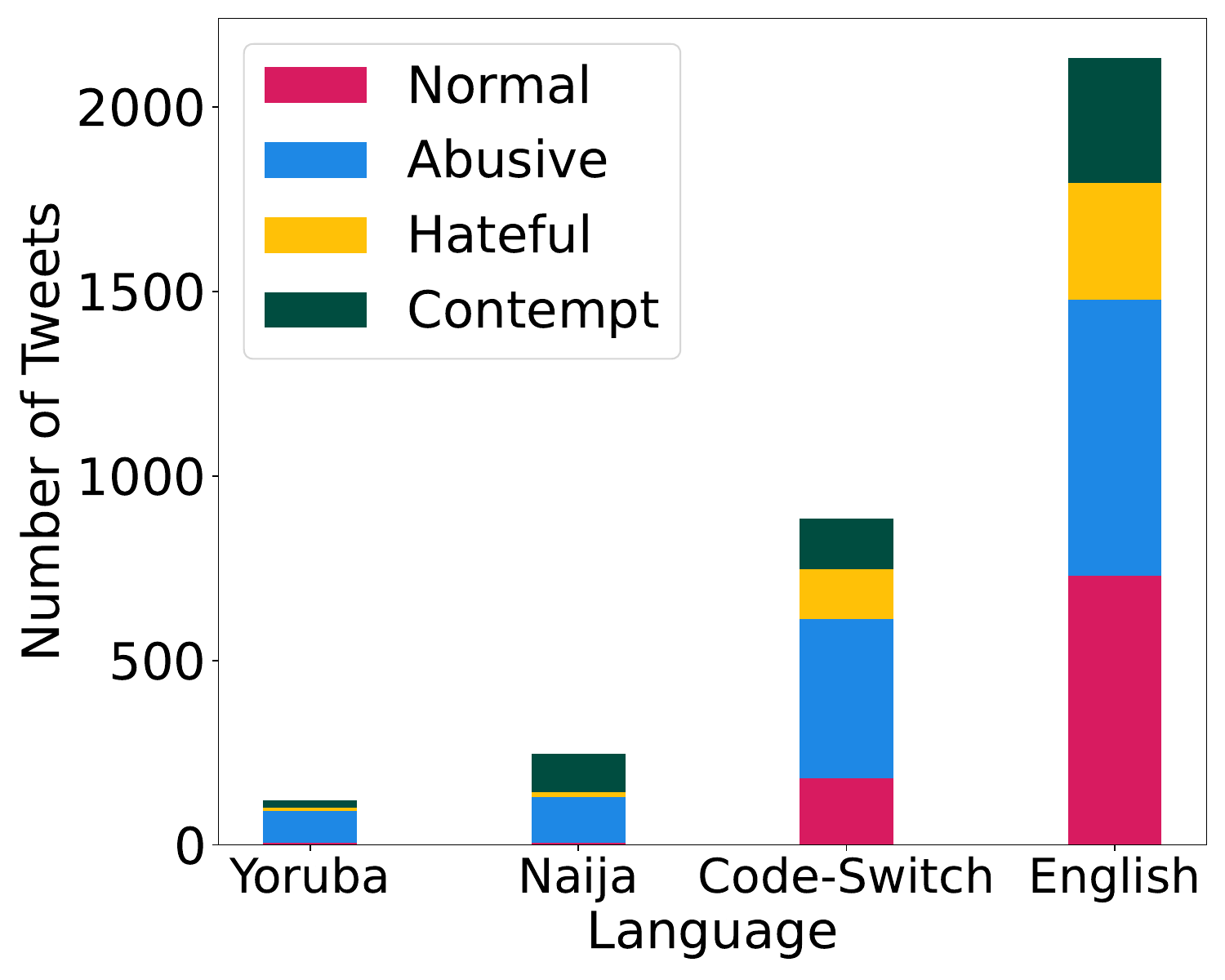}
    \vspace{-2mm}
    \caption{The label distribution according to languages.  \looseness-1}
    \label{fig:language_chart}
\end{figure}

\paragraph{\ekohate data statistics} \Cref{fig:candidate_chart}  shows the annotated data distribution for the three political candidates: Jandor, GRV, and Sanwo-Olu, with 332, 1385, and 1682 tweets respectively. The incumbent governor, representing APC, garnered the highest engagement, resulting in more tweets. %Across these candidates, tweets categorized as \textit{abusive}  accounted for $40\%$ of the total, while \textit{hateful} tweets were the least common category across the board.
%Across the candidates, \textit{abusive} tweets have a similar proportion ($41\%$), while for the \textit{hateful} label, tweets associated with GRV account are more than $4\%$ mor compared to other candidates. Similarly, \textit{contempt} label are more common for Jandor and GRV (i.e $\approx+8\%$) more than Sanwo-Olu. 
Among the candidates, the proportion of \textit{abusive} tweets is similar at $41\%$. In contrast, \textit{hateful} tweets associated with the GRV account exceed those from other candidates by more than $4\%$. Additionally, tweets with the \textit{contempt} are approximately $8\%$ more frequent for Jandor and GRV compared to Sanwo-Olu.

%label are more prevalent for Jandor and GRV, roughly $8\%$ more than Sanwo-Olu.

The dataset exhibits three primary characteristics: it is multilingual, features code-switching, and is inherently noisy due to its social media origin. It has tweets in English, Yoruba, and Nigerian Pidgin (or Naija), which are %among the 
commonly used languages in Nigeria. Moreover, it includes instances of code-switching between these languages. \Cref{fig:language_chart} shows the distribution of tweets across Yoruba, Naija, Code-Switch and English, with 120 ($3.5\%)$, 247 ($7.3\%$), 884 ($26.0\%$), and 2,144 ($63.2\%$)  tweets respectively. The \textit{abusive} tweets outnumber \textit{normal} tweets across all languages, with Yoruba, Code-Switch, and Naija tweets having a higher proportion of abusive content compared to other categories within each language.

%Out of the code-switched data was sourced from tweets directed at three musketeers of the governorship election of the most populous and economically vibrant state in Nigeria; Lagos state, with the view to detect offensive and hate speech on political discussion. Our annotation of the language use in our data shows that $63.1\%$ of the data is from English, while the others are either in Yoruba ($3.5\%$), Nigerian-Pidgin ($7.3\%$), or Code-switched ($26.0\%$) between English and the other two languages (Nigerian-Pidgin and Yoruba).

 We split the data per label into 70\%, 10\% and 20\% to create the training, development and test.  % split of the \textbf{EkoHate} dataset.  %, based on a three-hierarchical annotation scheme in other to effectively reconcile annotations given by the annotators. 

%code-switched data was sourced from tweets directed at three musketeers of the governorship election of the most populous and economically vibrant state in Nigeria; Lagos state, with the view to detect offensive and hate speech on political discussion. Our annotation of the language use in our data shows that $63.1\%$ of the data is from English, while the others are either in Yoruba ($3.5\%$), Nigerian-Pidgin ($7.3\%$), or Code-switched ($26.0\%$) between English and the other two languages (Nigerian-Pidgin and Yoruba).

% Dataset property 
%Noisy text u no get shame,did you remember you lost ur polling until .
\begin{table}[t]
\footnotesize
\begin{center}
\scalebox{0.9}{
\begin{tabular}{p{29mm}rrr}
\toprule
 &  \multicolumn{3}{c}{\textbf{Number of tweets}}  \\
\textbf{Data}  & \textbf{train} & \textbf{dev} & \textbf{test}  \\
\midrule
\multicolumn{4}{l}{\textbf{\textit{Binary }}}  \\
OLID (N-O) & $11,916$ & $1,324$ & $860$ \\
HateUS2020 (N-H) & $2,160$ & $240$ & $600$ \\
EkoHate (N-O) & $1,950$ & $278$ & $559$ \\
EkoHate (N-H) & $976$ & $139$ & $280$ \\
\midrule
\multicolumn{4}{l}{\textbf{\textit{Multi class }}}  \\
EkoHate (N-A-H) & $1,950$ & $278$ & $559$  \\
FountaHate (N-A-H) & $79,625$ & $2,042$ & $4,299$  \\
EkoHate (N-A-H-C) & $2,377$ & $339$ & $682$ \\
\bottomrule
\end{tabular}
 }
\end{center}
\vspace{-2mm}
\footnotesize
  \caption{The split of the different datasets  \looseness-1}
     \label{tab:data_split}
\end{table}

\begin{table*}[t]
 \begin{center}
 \footnotesize
 \scalebox{0.9}{
 \begin{tabular}{l|ccccc|c}
 \toprule
\textbf{schema} & \textbf{normal} & \textbf{offensive} & \textbf{abusive} & \textbf{hateful} & \textbf{contempt} & \textbf{F1} \\
\midrule
% SVM & & & & & \\
N-O & $93.4_{\pm 0.4}$ & $96.8_{\pm 0.2}$ & - & - & - & $95.1_{\pm 0.3}$ \\
N-H & $94.6_{\pm 0.3}$ & - & & $89.2_{\pm 0.7}$ & - &  $91.9_{\pm 0.5}$ \\ 
N-A-H & $93.4_{\pm 0.5}$ & - & $85.9_{\pm 1.3}$ & $55.4_{\pm 4.7}$ & - & $78.2_{\pm 2.2}$ \\
N-A-H-C & $90.5_{\pm 0.6}$ & - & $78.6_{\pm 0.8}$ & $51.1_{\pm 2.2}$  & $61.1_{\pm 1.7}$ & $70.3_{\pm 1.3}$ \\
\bottomrule
\end{tabular}
}
  \vspace{-2mm}
  \caption{Result of hateful and offensive language detection on EkoHate dataset.  \looseness-1}
  \label{tab:result1}
  \end{center}
\end{table*}

\begin{table*}[t]
 \begin{center}
 \footnotesize

 \begin{tabular}{lr}
 \begin{minipage}[t]{0.8\textwidth}
 \begin{center}
 \footnotesize
 %\begin{tabular}{l|ccc|c}
 \scalebox{0.9}{
 \begin{tabular}{p{2.4cm}|p{1.1cm}p{1.1cm}p{1.1cm}p{1.1cm}|p{1.1cm}}
 \toprule
\textbf{dataset} & \textbf{normal} & \textbf{offensive} & \textbf{abusive} & \textbf{hateful} & \textbf{F1} \\
\midrule
OLID  & $88.3_{\pm 0.2}$ & $69.5_{\pm 1.0}$ & - & - & $78.9_{\pm 0.6}$ \\
{    }$\rightarrow$ EkoHate & $69.2_{\pm 0.2}$ & $73.1_{\pm 0.4}$ & - & - & $71.1_{\pm 0.3}$ \\
EkoHate  & $93.4_{\pm 0.4}$ & $96.8_{\pm 0.2}$ & - & - & $95.1_{\pm 0.3}$ \\
{    }$\rightarrow$ OLID & $80.4_{\pm 0.7}$ & $63.2_{\pm 0.8}$ & - & - & $71.8_{\pm 0.7}$ \\
\midrule
HateUS2020 & $95.2_{\pm 0.5}$ & - & - & $60.7_{\pm 2.5}$ & $77.8_{\pm 1.5}$ \\
{    }$\rightarrow$ EkoHate & $83.1_{\pm 0.6}$ & - & - & $34.1_{\pm 4.7}$ & $58.6_{\pm 2.6}$ \\
EkoHate & $94.6_{\pm 0.3}$ & - & - & $89.2_{\pm 0.7}$ & $91.9_{\pm 0.5}$ \\
{    }$\rightarrow$ HateUS2020 & $87.2_{\pm 1.2}$ & - & - &  $38.3_{\pm 1.6}$ & $62.7_{\pm 1.4}$ \\
\midrule
FountaHate & $95.2_{\pm 0.1}$ & - & $89.0_{\pm 0.1}$ &  $41.1_{\pm 1.4}$ & $75.1_{\pm 0.5}$ \\
{    }$\rightarrow$ EkoHate & $63.5_{\pm 0.7}$ & - & $34.9_{\pm 2.7}$ &  $33.3_{\pm 2.3}$ & $43.9_{\pm 0.7}$ \\
EkoHate & $93.4_{\pm 0.5}$ & - & $85.9_{\pm 1.3}$ & $55.4_{\pm 4.7}$ & $78.2_{\pm 2.2}$ \\
{    }$\rightarrow$ FountaHate & $82.8_{\pm 0.7}$ & - & $61.2_{\pm 3.4}$ & $16.8_{\pm 1.5}$ & $53.6_{\pm 0.9}$ \\
\bottomrule
\end{tabular}
}
  \vspace{-2mm}
  \caption{Cross-corpus transfer results between EkoHate and other datasets.\looseness-1}
  \label{tab:result2}
  \end{center}
 \end{minipage}

\vspace{-2mm}
\end{tabular}

  \end{center}
\end{table*}

\section{Experiment Setup}

\begin{comment}
 & \begin{minipage}[t]{0.5\textwidth}
 \begin{center}
 \footnotesize
    \scalebox{0.97}{
    \setlength{\tabcolsep}{5.7pt}
    
    \begin{table*}[t]
    \begin{center}
    \footnotesize
    \begin{tabular}{p{13mm}|cccc}
    \toprule
     &  \multicolumn{4}{c}{\textbf{Language}}  \\
    \textbf{Data}  & \textbf{Eng.} & \textbf{CDW} & \textbf{Pcm.} & \textbf{Yor.}  \\
    \midrule
    N-O & $94.7_{\pm 0.3}$ & $95.4_{\pm 0.6}$ & $82.3_{\pm 0.0}$ & $100.0_{\pm 0.0}$ \\
    N-H & $91.7_{\pm 0.4}$ & $92.6_{\pm 0.8}$ & $73.3_{\pm 0.0}$ & $100.0_{\pm 0.2}$ \\
    N-A-H & $77.5_{\pm 0.6}$ & $78.0_{\pm 2.9}$ & $57.5_{\pm 7.0}$ & $91.4_{\pm 7.4}$ \\
    N-A-H-C & $68.9_{\pm 1.0}$ & $64.2_{\pm 2.7}$ & $60.4_{\pm 1.2}$ & $86.2_{\pm 12.7}$ \\
    \bottomrule
    \end{tabular}
    }
  \vspace{-1mm}
  \caption{In-language performance for English (Eng.), Code-Switch (CDW), Naija (Pcm.), and Yoruba (Yor.) on EkoHate testset.}
  \label{tab:lang_split_res}
  \end{center}
  \end{table*}
 \end{minipage}
\end{comment}

\paragraph{Dataset} For our study, we opted for both binary and multi-class settings. For binary settings with EkoHate, we consider \textbf{binary} label configurations: ``normal vs. offensive'' (N-O), and ``normal vs. hateful'' (N-H). For the multi-class, we consider: ``normal vs. abusive vs. hateful'' (N-A-H), and ``normal vs. abusive vs. hateful vs. contempt'' (N-A-H-C). And in the multi-class setup, we remove the instances of the excluded classes in the train, development and test splits.

To assess the quality and consistency of our annotations relative to previous work, we conducted cross-corpus transfer experiments. For this task, we opted for three widely known datasets which are offensive language identification dataset (OLID)~\citep{zampieri-etal-2019-predicting}, a corpus of offensive speech and stance detection from the 2020 US elections (HateUS2020)~\citep{grimminger-klinger-2021-hate}, and a large hatespeech dataset (FountaHate)~\citep{Founta_Djouvas_paper}. These are datasets collected from Twitter and manually annotated. While OLID used \textit{offensive} and \textit{non-offensive} schema, HateUS2020 used \textit{hateful} and \textit{non-hateful} schema, and FountaHate used four classes which are, \textit{normal}, \textit{abusive}, \textit{hateful}, and \textit{spam}. However, for this work, instances labeled as \textit{spam} were removed. 

OLID and HateUS2020 had no validation set, therefore, we sampled $10\%$ of their training splits as the development set. However, due to the large size of FountaHate and the absence of dedicated development and test sets, unlike OLID and HateUS2020, we split the data using the proportions of 92.5\%, 2.5\%, and 5\% for training, development, and test sets, respectively.
See \Cref{tab:data_split} for the splits and sizes of data.

\paragraph{Models and Training} Using the respective datasets, we fine-tuned Twitter-RoBERTa-base~\citep{barbieri-etal-2020-tweeteval}.~\footnote{While our data is multilingual and code-switched, we find that English-only model performed better than multilingual model from our early analysis. Result is in \autoref{sec:appendix}} Each model was trained for 10 epochs with a maximal input length  of 256, batch size of 16, a learning rate of $2 \cdot 10^{-5}$ using the Huggingface framework. We reported label-wise F1 score as well as  macro F1 of 5 runs for the different models for the different classes and also Macro-F1.

Furthermore, given that the baseline model was trained using 5 runs, we explored the effect of model ensembling on the EkoHate dataset. The use of model ensembling has been shown to achieve better results than individual models\citep{zimmerman-etal-2018-improving,rajendran-etal-2019-ubc,saha-etal-2021-hate,singhal-bedi-2024-transformers-lt}. Therefore, we also evaluated hard ensembling, which involved majority voting on five model predictions.

\begin{table*}[t]
 \begin{center}
 \footnotesize
\scalebox{0.87}{
 \begin{tabular}{l|p{8.5cm}|ccc}
 \toprule
\textbf{schema} & \textbf{Tweet} &Lang.& \textbf{Gold} & \textbf{Pred.} \\
\midrule
%\multirow{4}{*}{N-O} &
%We will kill the hoodlums disrupting this election process! it time to take law into our hands.  \\
%& Women belong to the kitchen and not in politics. \\
%& We hate small %boys, you are a small boy with no experience, you can’t rule us. \\
%& Leave that one to ur family members, nobody need ur bitter ass
%You are Igbo, you  can’t rule us in Lagos.  \\
%\midrule
%\multirow{4}{*}{N-H} &
%You are very stupid! \\
%& Olodo, oloriburuku \\
%& U be mumu , see gbadego ur mumu never do abi eke nparo funro. \\
%& Mumu your principal is using Eko o ni baje ...u r using Eko %edide..oloshi ..Ori yi ti o pe ye ma pe laipe.  \\
%\midrule
\multirow{4}{*}{N-A-H} &
Leave Lagos and return to Anambra omo werey & CDW & hateful & abusive   \\
& Ogun kill you!  By the time we're done with you, you'll tell us the real truth behind 20-10-2020. Murderer!  & CDW & hateful & abusive  \\
%& There's bomb in your brain. & hateful & abusive  \\
%& U go school so? Vapour abi wetin be ur name? \\
\midrule
{N-A-H-C} & The way pitobi failed you will also failed woefully & CDW & hateful & abusive \\
%& Your tribunal case is being prepared.  Enjoy the office while it lasts.  The actual election result is loading.  Your and your boss will be retired. & Eng. & hateful & contempt \\

%& Bro, go to the field and gather momentum. Social media can only do so much & normal & contempt \\
%& Thumb  to the working Governor! & normal & abusive \\
\bottomrule
\end{tabular}
}
  \vspace{-2mm}
  \caption{ Examples of correct and incorrect predictions.  \looseness-1}
  \label{tab:error}
  \end{center}
\end{table*}

 \begin{table}[t]
    \begin{center}
    \footnotesize
    \scalebox{0.8}{
    \begin{tabular}{p{13mm}|cccc}
    \toprule
     &  \multicolumn{4}{c}{\textbf{Language}}  \\
    \textbf{Data}  & \textbf{English} & \textbf{Code-Switch} & \textbf{Naija} & \textbf{Yoruba}  \\
    \midrule
    N-O & $94.7_{\pm 0.3}$ & $95.4_{\pm 0.6}$ & $82.3_{\pm 0.0}$ & $100.0_{\pm 0.0}$ \\
    N-H & $91.7_{\pm 0.4}$ & $92.6_{\pm 0.8}$ & $73.3_{\pm 0.0}$ & $100.0_{\pm 0.2}$ \\
    N-A-H & $77.5_{\pm 0.6}$ & $78.0_{\pm 2.9}$ & $57.5_{\pm 7.0}$ & $91.4_{\pm 7.4}$ \\
    N-A-H-C & $68.9_{\pm 1.0}$ & $64.2_{\pm 2.7}$ & $60.4_{\pm 1.2}$ & $86.2_{\pm 12.7}$ \\
    \bottomrule
    \end{tabular}
    }
  \vspace{-1mm}
  \caption{In-language performance for English, Code-Switch, Naija, and Yoruba on EkoHate test set.}
  \label{tab:lang_split_res}
  \end{center}
  \vspace{-2mm}
  \end{table}

\section{Results}
\paragraph{EkoHate baseline}
We fine-tuned Twitter-RoBERTa-base on the EkoHate dataset in both binary and multi-class settings and present the results in \Cref{tab:result1}. We observed that binary configurations are easy tasks, achieving high F1 scores of $95.1$ and 9$1.9$ for \textit{normal versus offensive and hateful} categories, respectively. However, multi-class configurations  are difficult, as classes are not predicted equally well. %Except for the N vs O, normal was well predicted for other configurations when compared to other classes. We also observed that the more classes we had N vs A vs H, and N vs A vs H vs, the more difficult it became with F1 less than $0.80$. 
Lastly, we observed that in all settings, the \textit{hateful} class was the most challenging. We attribute this to the \textit{hateful} class being the least occurring in the EkoHate dataset and the language model's inability to correctly model the class, despite being trained as few-shot learners. 
Due to class imbalance in the data, we explored models ensembling using majority voting. Our results indicate potential improvements of up to $+2.3$ for multi-class setups, with relative improvements observed in the binary setups. More details are provided in \Cref{sec:model_esembling}.

\paragraph{Effect of code-switching}
Going further, we examine the in-language performance of the baseline models, focusing on the F1 scores for the languages present in the test sets (English, Code-switch, Naija and Yoruba). %\Cref{tab:lang_split} 
\Cref{sec:testset_lang} shows the distribution of these languages in the test sets, while \Cref{tab:lang_split_res}  shows the corresponding results. The results indicate that the models struggle more with Naija, as shown by consistently lower average in-language performance compared to the overall test performance in \Cref{tab:result1}. We attribute this primarily to the small size of the Naija examples. In contrast, we observed higher F1 scores for Yoruba. However, considering both Yoruba and Naija have the fewest number of examples, we cautiously attribute their performances to chance and leave this for future work to explore.

\paragraph{Cross-corpus Transfer setting} For this experiment, we trained Twitter-RoBERTa-base on existing datasets and evaluated its performance on the EkoHate dataset and vice versa.  \Cref{tab:result2} shows the result of our zero-shot cross-corpus transfer result. As expected, when models trained on any of the datasets are evaluated on their corresponding test sets, we obtained a high F1 score with the lowest being FountaHate, where we obtained $75.1$ F1 score. However, when these models are evaluated on a different corpora, we observed significantly low performance, for example, HateUS2020$\rightarrow$EkoHate gave $58.6$ points. Surprisingly, transferring from our newly created data, EkoHate performs slightly better than OLID ($+1\%$) \& HateUS2020 ($+4\%$), which shows our dataset generalizes more, possibly due to the fact that EkoHate has a majority of English tweets. 

%Regarding the cross-corpus transfer comparison to other offensive speech datasets like Olid and HateUs2020 I dont think the results are very surprising. Their dataset contains over 60% English so it makes sense that models trained on their dataset would transfer decently to english datasets but that models trained purely english datasets would not transfer as well to their datasets.

% \begin{table*}[t]
%  \begin{center}
%  \footnotesize
% 
%  \begin{tabular}{l|c|cccc}
%  \toprule
% & \textbf{train size} & \textbf{normal} & \textbf{offensive} & \textbf{hateful} & \textbf{Mirco F1} \\
% \midrule
% OLID & $11,916$ & $88.3$ & $69.5$ & - & $78.9$ \\
% {  }$\rightarrow$ EkoHate & - & $69.2$ & $73.1$ & - & $71.1$ \\
% EkoHate & $1,950$  & $93.4$ & $96.8$ & - & $95.1$ \\
% {  }$\rightarrow$ OLID & - & $80.4$ & $63.2$ & - & $71.8$ \\
% \midrule
% HateUS2020 & $2,160$  & $95.2$ & - & $60.7$ & $77.8$ \\
% {  }$\rightarrow$ EkoHate & -  & $83.1$ & - & $34.1$ & $58.6$ \\
% EkoHate  & 975  & $94.6$ & - & $89.2$ & $91.9$ \\
% {  }$\rightarrow$ HateUS2020 & -  & $87.2$ & - & $38.3$ & $62.7$ \\
% \bottomrule
% \end{tabular}
% 
%   \vspace{-1mm}
%   \caption{Result of cross-corpus transfer for hate and offensive language identification. We report the average F1 of 5 runs for the different models for the different classes and also Micro-F1}
%   \label{tab:result2}
%   \end{center}
% \end{table*}

\section{Error Analysis}
Results from \Cref{tab:result1,tab:result2} show that the \textit{hateful} is a difficult class to correctly predict. Hence, we examined the predictions of one of the baseline models for the N-A-H and N-A-H-C. In \Cref{sec:cofus}, we showed that \textit{hateful} tweets were often misclassified as \textit{abusive}. \Cref{tab:error} highlights some misclassified \textit{hateful} tweets. For example, the first N-A-H example expressed hatred toward someone who perhaps is non-Lagosian, asking them to return to their place of origin (\textit{Anambra}) after referring to them as a \textit{mad person} (\textit{omo werey}). The second example is a wish for the recipient to be killed by \textit{Ogun}\footnote{Yoruba god of iron and war.}, while the third example shows the recipient being wished failure just like Pitobi (Peter Obi\footnote{Nigeria's LP presidential candidate in the 2023 elections.}). However, the models failed to capture these tweets as \textit{hateful}. See \Cref{tab:error2} for more examples.

\section{Related Work}
Several works have been conducted to create hate speech datasets, but the majority have focused on English and other high-resource languages, often within the context of specific countries~\citep{Mathew_Saha_Yimam_2021, demus-etal-2022-comprehensive, ron-etal-2023-factoring, ayele-etal-2023-multilingual, piot2024metahate}. However, in the context of Africa, only a few hate speech datasets exist to the best of our knowledge. For example, ~\citet{ayele-etal-2023-exploring} created a hate speech dataset for Amharic tweets using a hate and non-hate speech schema, while ~\citet{aliyu2022herdphobia} created a dataset for detecting hate speech against Fulani herders using hate/non-hate/indeterminate schema. These works, however, primarily focused on racial hate. In this work, we focused on election-related hate speech, which includes racial elements.

\section{Conclusion}
In this paper, we present \textbf{EkoHate} dataset for offensive and hate speech detection. Our dataset is code-switched and focused on political discussion in the last 2023 Lagos elections. We conducted empirical evaluations in fully supervised settings, covering both binary and multi-class tasks, finding multi-class to be more challenging. However, ensemble methods slightly improved multi-class performance. Additionally, cross-corpus experiments between EkoHate and existing datasets confirmed our annotations' alignment and our dataset's usefulness. 

%We conduct empirical evaluations in a fully supervised setting, including both binary and multi-class classification tasks. Our results indicate that the multi-class setup presents more challenges. However, we demonstrate that ensemble methods can slightly improve performance in the multi-class setting. Lastly, we perform cross-corpus transfer experiments between EkoHate and existing datasets, showing that our annotations align with these datasets, which confirms the usefulness of our dataset.
%We conduct empirical evaluation in both full-supervised and cross-corpus transfers which confirms the usefulness of our dataset. 

%\section{Bib\TeX{} Files}
%\label{sec:bibtex}

%\newpage
\subsubsection*{Acknowledgments}
Jesujoba Alabi was partially supported by the BMBF’s (German Federal Ministry of Education and Research) SLIK project under the grant 01IS22015C. David Adelani acknowledges the support of DeepMind Academic Fellowship programme. Oluwaseyi Adeyemo acknowledges the support of the Founder, Afe Babalola University, Ado-Ekiti, Nigeria. Lastly, we thank Feyisayo Olalere, Nicholas Howell, the anonymous reviewers of AfricaNLP 2024 workshop and WOAH 2024 for their helpful feedback.

\bibliography{anthology,custom}

\begin{thebibliography}{31}
\expandafter\ifx\csname natexlab\endcsname\relax\def\natexlab#1{#1}\fi

\bibitem[{Alabi et~al.(2022)Alabi, Adelani, Mosbach, and Klakow}]{alabi-etal-2022-adapting}
Jesujoba~O. Alabi, David~Ifeoluwa Adelani, Marius Mosbach, and Dietrich Klakow. 2022.
\newblock \href {https://aclanthology.org/2022.coling-1.382} {Adapting pre-trained language models to {A}frican languages via multilingual adaptive fine-tuning}.
\newblock In \emph{Proceedings of the 29th International Conference on Computational Linguistics}, pages 4336--4349, Gyeongju, Republic of Korea. International Committee on Computational Linguistics.

\bibitem[{Alfina et~al.(2017)Alfina, Mulia, Fanany, and Ekanata}]{Alfina_2017}
Ika Alfina, Rio Mulia, Mohamad~Ivan Fanany, and Yudo Ekanata. 2017.
\newblock \href {https://doi.org/10.1109/ICACSIS.2017.8355039} {Hate speech detection in the indonesian language: A dataset and preliminary study}.
\newblock In \emph{2017 International Conference on Advanced Computer Science and Information Systems (ICACSIS)}, pages 233--238.

\bibitem[{Aliyu et~al.(2022)Aliyu, Wajiga, Murtala, Muhammad, Abdulmumin, and Ahmad}]{aliyu2022herdphobia}
Saminu~Mohammad Aliyu, Gregory~Maksha Wajiga, Muhammad Murtala, Shamsuddeen~Hassan Muhammad, Idris Abdulmumin, and Ibrahim~Said Ahmad. 2022.
\newblock \href {http://arxiv.org/abs/2211.15262} {Herdphobia: A dataset for hate speech against fulani in nigeria}.

\bibitem[{Ayele et~al.(2023{\natexlab{a}})Ayele, Dinter, Yimam, and Biemann}]{ayele-etal-2023-multilingual}
Abinew~Ali Ayele, Skadi Dinter, Seid~Muhie Yimam, and Chris Biemann. 2023{\natexlab{a}}.
\newblock \href {https://aclanthology.org/2023.ranlp-1.5} {Multilingual racial hate speech detection using transfer learning}.
\newblock In \emph{Proceedings of the 14th International Conference on Recent Advances in Natural Language Processing}, pages 41--48, Varna, Bulgaria. INCOMA Ltd., Shoumen, Bulgaria.

\bibitem[{Ayele et~al.(2023{\natexlab{b}})Ayele, Yimam, Belay, Asfaw, and Biemann}]{ayele-etal-2023-exploring}
Abinew~Ali Ayele, Seid~Muhie Yimam, Tadesse~Destaw Belay, Tesfa Asfaw, and Chris Biemann. 2023{\natexlab{b}}.
\newblock \href {https://aclanthology.org/2023.ranlp-1.6} {Exploring {A}mharic hate speech data collection and classification approaches}.
\newblock In \emph{Proceedings of the 14th International Conference on Recent Advances in Natural Language Processing}, pages 49--59, Varna, Bulgaria. INCOMA Ltd., Shoumen, Bulgaria.

\bibitem[{Barbieri et~al.(2020)Barbieri, Camacho-Collados, Espinosa~Anke, and Neves}]{barbieri-etal-2020-tweeteval}
Francesco Barbieri, Jose Camacho-Collados, Luis Espinosa~Anke, and Leonardo Neves. 2020.
\newblock \href {https://doi.org/10.18653/v1/2020.findings-emnlp.148} {{T}weet{E}val: Unified benchmark and comparative evaluation for tweet classification}.
\newblock In \emph{Findings of the Association for Computational Linguistics: EMNLP 2020}, pages 1644--1650, Online. Association for Computational Linguistics.

\bibitem[{Carlson(2020)}]{matt_carlson_2020}
Matt Carlson. 2020.
\newblock \href {https://doi.org/10.1080/1369118X.2018.1505934} {Fake news as an informational moral panic: the symbolic deviancy of social media during the 2016 us presidential election}.
\newblock \emph{Information, Communication \& Society}, 23(3):374--388.

\bibitem[{Carney(2022)}]{carney2022effect}
Kevin Carney. 2022.
\newblock The effect of social media on voters: experimental evidence from an indian election.
\newblock \emph{Job Market Paper}, pages 1--44.

\bibitem[{Conneau et~al.(2019)Conneau, Khandelwal, Goyal, Chaudhary, Wenzek, Guzm{\'{a}}n, Grave, Ott, Zettlemoyer, and Stoyanov}]{xlm_roberta_Conneau}
Alexis Conneau, Kartikay Khandelwal, Naman Goyal, Vishrav Chaudhary, Guillaume Wenzek, Francisco Guzm{\'{a}}n, Edouard Grave, Myle Ott, Luke Zettlemoyer, and Veselin Stoyanov. 2019.
\newblock \href {http://arxiv.org/abs/1911.02116} {Unsupervised cross-lingual representation learning at scale}.
\newblock \emph{CoRR}, abs/1911.02116.

\bibitem[{Davidson et~al.(2017)Davidson, Warmsley, Macy, and Weber}]{hateoffensive}
Thomas Davidson, Dana Warmsley, Michael Macy, and Ingmar Weber. 2017.
\newblock Automated hate speech detection and the problem of offensive language.
\newblock In \emph{Proceedings of the 11th International AAAI Conference on Web and Social Media}, ICWSM '17, pages 512--515.

\bibitem[{Demus et~al.(2022)Demus, Pitz, Sch{\"u}tz, Probol, Siegel, and Labudde}]{demus-etal-2022-comprehensive}
Christoph Demus, Jonas Pitz, Mina Sch{\"u}tz, Nadine Probol, Melanie Siegel, and Dirk Labudde. 2022.
\newblock \href {https://doi.org/10.18653/v1/2022.woah-1.14} {Detox: A comprehensive dataset for {G}erman offensive language and conversation analysis}.
\newblock In \emph{Proceedings of the Sixth Workshop on Online Abuse and Harms (WOAH)}, pages 143--153, Seattle, Washington (Hybrid). Association for Computational Linguistics.

\bibitem[{Febriana and Budiarto(2019)}]{Febriana_2019}
Trisna Febriana and Arif Budiarto. 2019.
\newblock \href {https://doi.org/10.1109/ICIMTech.2019.8843722} {Twitter dataset for hate speech and cyberbullying detection in indonesian language}.
\newblock In \emph{2019 International Conference on Information Management and Technology (ICIMTech)}, volume~1, pages 379--382.

\bibitem[{Founta et~al.(2018)Founta, Djouvas, Chatzakou, Leontiadis, Blackburn, Stringhini, Vakali, Sirivianos, and Kourtellis}]{Founta_Djouvas_paper}
Antigoni Founta, Constantinos Djouvas, Despoina Chatzakou, Ilias Leontiadis, Jeremy Blackburn, Gianluca Stringhini, Athena Vakali, Michael Sirivianos, and Nicolas Kourtellis. 2018.
\newblock \href {https://doi.org/10.1609/icwsm.v12i1.14991} {Large scale crowdsourcing and characterization of twitter abusive behavior}.
\newblock \emph{Proceedings of the International AAAI Conference on Web and Social Media}, 12(1).

\bibitem[{Fujiwara et~al.(2021)Fujiwara, Müller, and Schwarz}]{NBERw28849}
Thomas Fujiwara, Karsten Müller, and Carlo Schwarz. 2021.
\newblock \href {https://doi.org/10.3386/w28849} {The effect of social media on elections: Evidence from the united states}.
\newblock Working Paper 28849, National Bureau of Economic Research.

\bibitem[{Grimminger and Klinger(2021)}]{grimminger-klinger-2021-hate}
Lara Grimminger and Roman Klinger. 2021.
\newblock \href {https://aclanthology.org/2021.wassa-1.18} {Hate towards the political opponent: A {T}witter corpus study of the 2020 {US} elections on the basis of offensive speech and stance detection}.
\newblock In \emph{Proceedings of the Eleventh Workshop on Computational Approaches to Subjectivity, Sentiment and Social Media Analysis}, pages 171--180, Online. Association for Computational Linguistics.

\bibitem[{Grinberg et~al.(2019)Grinberg, Joseph, Friedland, Swire-Thompson, and Lazer}]{Grinberg_2019}
Nir Grinberg, Kenneth Joseph, Lisa Friedland, Briony Swire-Thompson, and David Lazer. 2019.
\newblock \href {https://doi.org/10.1126/science.aau2706} {Fake news on twitter during the 2016 u.s. presidential election}.
\newblock \emph{Science}, 363(6425):374--378.

\bibitem[{Liu et~al.(2019)Liu, Ott, Goyal, Du, Joshi, Chen, Levy, Lewis, Zettlemoyer, and Stoyanov}]{liu2019roberta}
Yinhan Liu, Myle Ott, Naman Goyal, Jingfei Du, Mandar Joshi, Danqi Chen, Omer Levy, Mike Lewis, Luke Zettlemoyer, and Veselin Stoyanov. 2019.
\newblock Roberta: A robustly optimized bert pretraining approach.
\newblock \emph{arXiv preprint arXiv:1907.11692}.

\bibitem[{Mathew et~al.(2021)Mathew, Saha, Yimam, Biemann, Goyal, and Mukherjee}]{Mathew_Saha_Yimam_2021}
Binny Mathew, Punyajoy Saha, Seid~Muhie Yimam, Chris Biemann, Pawan Goyal, and Animesh Mukherjee. 2021.
\newblock \href {https://doi.org/10.1609/aaai.v35i17.17745} {Hatexplain: A benchmark dataset for explainable hate speech detection}.
\newblock \emph{Proceedings of the AAAI Conference on Artificial Intelligence}, 35(17):14867--14875.

\bibitem[{Nwozor et~al.(2022)Nwozor, Ajakaiye, Okidu, Olanrewaju, and Afolabi}]{nwozor2022social}
Agaptus Nwozor, Olanrewaju~OP Ajakaiye, Onjefu Okidu, Alex Olanrewaju, and Oladiran Afolabi. 2022.
\newblock Social media in politics: Interrogating electorate-driven hate speech in nigeria's 2019 presidential campaigns.
\newblock \emph{JeDEM-eJournal of eDemocracy and Open Government}, 14(1):104--129.

\bibitem[{Piot et~al.(2024)Piot, Martín-Rodilla, and Parapar}]{piot2024metahate}
Paloma Piot, Patricia Martín-Rodilla, and Javier Parapar. 2024.
\newblock \href {http://arxiv.org/abs/2401.06526} {Metahate: A dataset for unifying efforts on hate speech detection}.

\bibitem[{Rajendran et~al.(2019)Rajendran, Zhang, and Abdul-Mageed}]{rajendran-etal-2019-ubc}
Arun Rajendran, Chiyu Zhang, and Muhammad Abdul-Mageed. 2019.
\newblock \href {https://doi.org/10.18653/v1/S19-2136} {{UBC}-{NLP} at {S}em{E}val-2019 task 6: Ensemble learning of offensive content with enhanced training data}.
\newblock In \emph{Proceedings of the 13th International Workshop on Semantic Evaluation}, pages 775--781, Minneapolis, Minnesota, USA. Association for Computational Linguistics.

\bibitem[{Ron et~al.(2023)Ron, Levi, Oshri, and Shenhav}]{ron-etal-2023-factoring}
Gal Ron, Effi Levi, Odelia Oshri, and Shaul Shenhav. 2023.
\newblock \href {https://doi.org/10.18653/v1/2023.woah-1.21} {Factoring hate speech: A new annotation framework to study hate speech in social media}.
\newblock In \emph{The 7th Workshop on Online Abuse and Harms (WOAH)}, pages 215--220, Toronto, Canada. Association for Computational Linguistics.

\bibitem[{Saha et~al.(2021)Saha, Paharia, Chakraborty, Saha, and Mukherjee}]{saha-etal-2021-hate}
Debjoy Saha, Naman Paharia, Debajit Chakraborty, Punyajoy Saha, and Animesh Mukherjee. 2021.
\newblock \href {https://aclanthology.org/2021.dravidianlangtech-1.38} {Hate-alert@{D}ravidian{L}ang{T}ech-{EACL}2021: Ensembling strategies for transformer-based offensive language detection}.
\newblock In \emph{Proceedings of the First Workshop on Speech and Language Technologies for Dravidian Languages}, pages 270--276, Kyiv. Association for Computational Linguistics.

\bibitem[{Sasu(2023)}]{doris_doris_sasu_stat}
Doris~Dokua Sasu. 2023.
\newblock \href {https://www.statista.com/statistics/1308467/population-of-lagos-nigeria/} {Population of lagos, nigeria 2000-2035}.
\newblock \emph{statista}.

\bibitem[{Siegel et~al.(2021)Siegel, Nikitin, Barberá, Sterling, Pullen, Bonneau, Nagler, and Tucker}]{QJPS_19045}
Alexandra~A. Siegel, Evgenii Nikitin, Pablo Barberá, Joanna Sterling, Bethany Pullen, Richard Bonneau, Jonathan Nagler, and Joshua~A. Tucker. 2021.
\newblock \href {https://doi.org/10.1561/100.00019045} {Trumping hate on twitter? online hate speech in the 2016 u.s. election campaign and its aftermath}.
\newblock \emph{Quarterly Journal of Political Science}, 16(1):71--104.

\bibitem[{Singhal and Bedi(2024)}]{singhal-bedi-2024-transformers-lt}
Kriti Singhal and Jatin Bedi. 2024.
\newblock \href {https://aclanthology.org/2024.ltedi-1.32} {Transformers@{LT}-{EDI}-{EACL}2024: Caste and migration hate speech detection in {T}amil using ensembling on transformers}.
\newblock In \emph{Proceedings of the Fourth Workshop on Language Technology for Equality, Diversity, Inclusion}, pages 249--253, St. Julian's, Malta. Association for Computational Linguistics.

\bibitem[{Suryawanshi et~al.(2020)Suryawanshi, Chakravarthi, Arcan, and Buitelaar}]{suryawanshi-etal-2020-multimodal}
Shardul Suryawanshi, Bharathi~Raja Chakravarthi, Mihael Arcan, and Paul Buitelaar. 2020.
\newblock \href {https://aclanthology.org/2020.trac-1.6} {Multimodal meme dataset ({M}ulti{OFF}) for identifying offensive content in image and text}.
\newblock In \emph{Proceedings of the Second Workshop on Trolling, Aggression and Cyberbullying}, pages 32--41, Marseille, France. European Language Resources Association (ELRA).

\bibitem[{Yerlikaya and Toker(2020)}]{turgay_2020}
Turgay Yerlikaya and Seca Toker. 2020.
\newblock \href {https://api.semanticscholar.org/CorpusID:225728790} {Social media and fake news in the post-truth era: The manipulation of politics in the election process}.
\newblock \emph{Insight Turkey}, 22:177--196.

\bibitem[{Zahrah et~al.(2022)Zahrah, Nurse, and Goldsmith}]{Zahrah_2022}
Fatima Zahrah, Jason R.~C. Nurse, and Michael Goldsmith. 2022.
\newblock \href {https://doi.org/10.1145/3477314.3507226} {A comparison of online hate on reddit and 4chan: a case study of the 2020 us election}.
\newblock In \emph{Proceedings of the 37th ACM/SIGAPP Symposium on Applied Computing}, SAC '22, page 1797–1800, New York, NY, USA. Association for Computing Machinery.

\bibitem[{Zampieri et~al.(2019)Zampieri, Malmasi, Nakov, Rosenthal, Farra, and Kumar}]{zampieri-etal-2019-predicting}
Marcos Zampieri, Shervin Malmasi, Preslav Nakov, Sara Rosenthal, Noura Farra, and Ritesh Kumar. 2019.
\newblock \href {https://doi.org/10.18653/v1/N19-1144} {Predicting the type and target of offensive posts in social media}.
\newblock In \emph{Proceedings of the 2019 Conference of the North {A}merican Chapter of the Association for Computational Linguistics: Human Language Technologies, Volume 1 (Long and Short Papers)}, pages 1415--1420, Minneapolis, Minnesota. Association for Computational Linguistics.

\bibitem[{Zimmerman et~al.(2018)Zimmerman, Kruschwitz, and Fox}]{zimmerman-etal-2018-improving}
Steven Zimmerman, Udo Kruschwitz, and Chris Fox. 2018.
\newblock \href {https://aclanthology.org/L18-1404} {Improving hate speech detection with deep learning ensembles}.
\newblock In \emph{Proceedings of the Eleventh International Conference on Language Resources and Evaluation ({LREC} 2018)}, Miyazaki, Japan. European Language Resources Association (ELRA).

\end{thebibliography}
%\bibliography{custom}

\appendix

\section{Performance using different pre-trained language models}
\label{sec:appendix}

We compared the performance of RoBERTa~\cite{liu2019roberta} (English PLM model), XLM-RoBERTa~\cite{xlm_roberta_Conneau} (multilingual PLM trained on 100 languages excluding Nigerian-Pidgin and Yoruba), Twitter-RoBERTa~\citep{barbieri-etal-2020-tweeteval} (trained on English tweets) and AfroXLMR~\citep{alabi-etal-2022-adapting} (an African-centric PLM that cover English, Nigerian-Pidgin, and Yoruba in it's pre-training).  Our results show that the English models have better performance than the multilingual variants, and the Twitter domain PLM have a similar performance as the RoBERTa model trained on the general domain. We have decided to use the Twitter domain-specific model for the remaining experiments. 
 \begin{table}[h]
\footnotesize
\begin{center}
%\scalebox{0.8}{
\begin{tabular}{p{60mm}r}
\toprule
{\textbf{Models}} &  {\textbf{F1}}  \\

\midrule
RoBERTa-base~\citep{liu2019roberta} & $70.4_{\pm 1.2}$ \\
XLM-RoBERTa-base~\citep{xlm_roberta_Conneau} & $66.5_{\pm 1.5}$  \\
Twitter-RoBERTa-base~\citep{barbieri-etal-2020-tweeteval} & $70.3_{\pm 1.1}$ \\
AfroXLM-RoBERTa-base~\citep{alabi-etal-2022-adapting} & $69.9_{\pm 1.0}$  \\
\bottomrule
\end{tabular}
% }
\end{center}
\footnotesize
  \caption{Comparing variants of RoBERTa on EkoHate N-A-H-C. We report the average Macro F1 after 5 runs.   \looseness-1}
     \label{tab:model_compare}
\end{table}

\section{Languages in the test sets of EkoHate} 
\label{sec:testset_lang}
EkoHate contains tweets in English, Yoruba, Naija, and their code-switched versions. While \Cref{fig:language_chart} provides a plot comparing the distribution of these languages in the whole dataset, \Cref{tab:lang_split} shows the distribution of these languages within the test split of each EkoHate schema. Yoruba and Naija have the smallest proportion in the test sets.

\begin{table}[h]
\footnotesize
\begin{center}
\scalebox{0.9}{
\begin{tabular}{p{18mm}|cccc}
\toprule
 &  \multicolumn{4}{c}{\textbf{Number of tweets}}  \\
\textbf{Data}  & \textbf{English} & \textbf{Code-Switch} & \textbf{Naija} & \textbf{Yoruba}  \\
\midrule
N-O & $364$ & $150$ & $25$ & $20$  \\
N-H & $212$ & $62$ & $4$ & $2$ \\
N-A-H & $364$ & $150$ & $25$ & $20$  \\
N-A-H-C & $437$ & $170$ & $49$ & $26$  \\
\bottomrule
\end{tabular}
}
\end{center}
\footnotesize
  \caption{Language distribution in the EkoHate test sets for English, Code-Switch, Naija, and Yoruba.  \looseness-1}
     \label{tab:lang_split}
\end{table}

\section{Error analysis with confusion matrix} 
\label{sec:cofus}

\Cref{tab:result1,tab:result2} shows that the different models struggle with correctly classifying the hateful class. Hence, we examined the predictions of the baseline models in the multi-class setup by computing the confusion matrices for the N-A-H and N-A-H-C, as presented in \Cref{tab:conf_1,tab:conf_2}, respectively, comparing the counts of correct and incorrect predictions given the ground truth and the predictions.

\begin{table}[ht]
\footnotesize
\begin{center}
%\scalebox{0.8}{
\begin{tabular}{ll|ccc|c}
\toprule
 & & \multicolumn{4}{c}{\textbf{Prediction}}  \\
& & \textbf{normal} & \textbf{abusive} & \textbf{hateful} & \textbf{Total}  \\
\midrule
\multirow{4}{*}{\rotatebox{90}{\textbf{Gold}}} & 
\textbf{normal} & $173$ & $5$ & $7$ & $185$  \\
& \textbf{abusive} & $5$ & $236$ & $38$ & $279$ \\
& \textbf{hateful} & $8$ & $38$ & $49$ & $95$  \\
\midrule
& \textbf{Total} & $186$ & $279$ & $94$ & $559$  \\
\bottomrule
\end{tabular}
% }
\end{center}
\footnotesize
  \caption{Confusion Matrix of one of the models trained and evaluated on EkoHate N-A-H.  \looseness-1}
     \label{tab:conf_1}
\end{table}

\Cref{tab:conf_1} shows that the baseline model struggle with classifying between abusive and \textit{hateful} tweets in the N-A-H setup, where $40\%$ of \textit{hateful} tweets were misclassified as \textit{abusive}, while $13.5\%$ of \textit{abusive} tweets were predicted as \textit{hateful}. With the inclusion of \textit{contempt} in the label schema, as we have in N-A-H-C, \Cref{tab:conf_2} shows that more \textit{abusive} tweets were classified as \textit{contempt} than as hateful, with $12.9\%$ and $7.5\%$, respectively. However, $36.8\%$ of hateful tweets were misclassified as abusive, showing how difficult it is for the models to correctly classify hateful tweets which forms the smallest portion of EkoHate.

\begin{table}[t]
\footnotesize
\begin{center}
\scalebox{0.8}{
\begin{tabular}{ll|cccc|c}
\toprule
 & & \multicolumn{4}{c}{\textbf{Prediction}}  \\
& & \textbf{normal} & \textbf{abusive} & \textbf{hateful} & \textbf{contempt} & \textbf{Total}  \\
\midrule
\multirow{4}{*}{\rotatebox{90}{\textbf{Gold}}} & 
\textbf{normal} & $166$ & $4$ & $2$ & $13$ & $185$ \\
& \textbf{abusive} & $2$ & $220$ & $21$ & $36$ & $279$ \\
& \textbf{hateful} & $5$ & $35$ & $42$ & $13$ & $95$  \\
& \textbf{contempt} & $11$ & $30$ & $6$ & $76$ & $123$ \\
\midrule
& \textbf{Total} & $184$ & $289$ & $71$ & $138$ & $682$  \\
\bottomrule
\end{tabular}
}
\end{center}
\footnotesize
  \caption{Confusion Matrix of one of the models trained and evaluated on EkoHate N-A-H-C.  \looseness-1}
     \label{tab:conf_2}
\end{table}

\section{Effect of model ensembling} 
\label{sec:model_esembling}
Given the result of the baseline model, we investigate the use of model ensembling,  which has been shown to improve model performance by leveraging the different strengths of various underlying models in class imbalance setups like ours. Therefore, instead of reporting the average F1 score, we opted to assess the impact of ensembling the 5 runs of the EkoHate baseline models. \Cref{tab:ensemb} shows a $+0.6$ improvement in the N-A-H and $+2.3$ improvement in the N-A-H-C scheme with ensembling, while binary schemes showed only marginal improvement, perhaps due to their initially good performance. We leave further analysis with model ensembling for future work.

\begin{table}[t]
\footnotesize
\begin{center}
\scalebox{0.99}{
 \begin{tabular}{p{1.5cm}|p{1.0cm}}
 \toprule
\textbf{schema} & \textbf{F1} \\
\midrule
N-O  &  $95.3$ \\
N-H  & $92.0$ \\
N-A-H  &  $78.8$ \\
N-A-H-C &  $72.3$ \\
\bottomrule
\end{tabular}
}
\end{center}
\footnotesize
  \caption{Model ensembling results on EkoHate dataset.  \looseness-1}
     \label{tab:ensemb}
\end{table}

\section{Annotation guidelines for EkoHate} 

\paragraph{Introduction}
This document presents guidelines on how to annotate \underline{potentially harmful tweets} that can cause emotional distress to individuals, incite violence, or discriminate against, and exclude social groups. 

As an annotator, it is important to approach this task with objectivity (as much as possible). We welcome your feedback on how we can update the guidelines based on the peculiarity of the language you are annotating, your background, or any socio-linguistic knowledge that we may have overlooked. Consider the following when performing the task:

Always use the guidelines and you should be objective and consistent in your annotation. 

\begin{itemize}
  \item \underline{Focus on the message conveyed} in the tweets and try not to focus on your personal opinion on the topic. 
  \item Do not rush to finish the task and always reach out to your language coordinator with  questions when in doubt.
\end{itemize}

\paragraph{Mental health risk and well-being}
Annotating harmful content can be psychologically distressing. We advise any annotator who feels anxious or uncomfortable during the process to take a break or stop the task and seek help. Early intervention is the best way to cope.

\paragraph{Definitions}

\begin{itemize}
\item  \textbf{Hate speech} is language content that expresses hatred towards a particular \textbf{group or individual} based on their political affiliation, race, ethnicity, religion, gender, sexual orientation, or other characteristics. \textbf{It also includes threats of violence}.

\item \textbf{Abusive language} is any form of bad language expressions including rude, impolite, insulting or belittling utterance intended to offend or harm an individual.

\item \textbf{Contempt} is any form of language that \textbf{conveys a strong disliking of, or negative attitudes} towards a targeted individual or group, and does so in a \textbf{neutral tone} or form of expression.
%\item Prejudice the expression of negative thoughts/beliefs regarding a targeted group or individual on the basis of the group’s characteristics, and/or (negative) monolithic references to the targeted group.
\item  \textbf{Indeterminate} is any tweet that is not \textbf{readable} or is \textbf{completely} written in another language other than your language of annotation. 
\item \textbf{Normal} is any form of expression  that does not contain any bad language belonging to any of the above classifications.
\end{itemize}

\paragraph{Task} Given a tweet, select the option that best describes it. \Cref{tab:sample} show examples of tweets classified as hate, offensive, contempt, intermediate, and normal.

\begin{table*}[t]
 \begin{center}
 \footnotesize

 \begin{tabular}{l|l}
 \toprule
\textbf{Label} & \textbf{Tweet} \\
\midrule
\multirow{4}{*}{Hateful} &
We will kill the hoodlums disrupting this election process! it time to take law into our hands.  \\
& Women belong to the kitchen and not in politics. \\
& We hate small boys, you are a small boy with no experience, you can’t rule us. \\
& Leave that one to ur family members, nobody need ur bitter ass
You are Igbo, you  can’t rule us in Lagos.  \\
\midrule
\multirow{4}{*}{Abusive} &
You are very stupid! \\
& Olodo, oloriburuku \\
& U be mumu , see gbadego ur mumu never do abi eke nparo funro. \\
& Mumu your principal is using Eko o ni baje ...u r using Eko edide..oloshi ..Ori yi ti o pe ye ma pe laipe.  \\
\midrule
\multirow{4}{*}{Contempt} &
Joker  \\
& Dide Go Where  \\
& Just dey play oooo  \\
& U go school so? Vapour abi wetin be ur name? \\
\midrule
\multirow{4}{*}{Normal} &
I will vote for you. \\
& My Incoming Governor. \\
& Godbless you \\
& May his soul rest in peace \\
\midrule
\multirow{2}{*}{Indeterminate} &
Tweets that are completely written in languages other than English and Nigerian  language of annotation. \\
& Tweets that make no sense or do not have any meaning \\
\bottomrule
\end{tabular}

  \vspace{-1mm}
  \caption{Examples of tweets classified as hateful, abusive, contempt, intermediate, and normal.  \looseness-1}
  \label{tab:sample}
  \end{center}
\end{table*}

\begin{table*}[t]
 \begin{center}
 \footnotesize

 \begin{tabular}{l|p{8.5cm}|ccc}
 \toprule
\textbf{schema} & \textbf{Tweet} &Lang.& \textbf{Gold} & \textbf{Pred.} \\
\midrule
%\multirow{4}{*}{N-O} &
%We will kill the hoodlums disrupting this election process! it time to take law into our hands.  \\
%& Women belong to the kitchen and not in politics. \\
%& We hate small %boys, you are a small boy with no experience, you can’t rule us. \\
%& Leave that one to ur family members, nobody need ur bitter ass
%You are Igbo, you  can’t rule us in Lagos.  \\
%\midrule
%\multirow{4}{*}{N-H} &
%You are very stupid! \\
%& Olodo, oloriburuku \\
%& U be mumu , see gbadego ur mumu never do abi eke nparo funro. \\
%& Mumu your principal is using Eko o ni baje ...u r using Eko %edide..oloshi ..Ori yi ti o pe ye ma pe laipe.  \\
%\midrule
\multirow{4}{*}{N-A-H} &
Leave Lagos and return to Anambra omo werey & CDW & hateful & abusive   \\
& Ogun kill you!  By the time we're done with you, you'll tell us the real truth behind 20-10-2020. Murderer!  & CDW & hateful & abusive  \\
& There's bomb in your brain. & Eng. & hateful & abusive  \\
%&& U go school so? Vapour abi wetin be ur name? \\
\midrule
\multirow{4}{*}{N-A-H-C} 
& Your tribunal case is being prepared.  Enjoy the office while it lasts.  The actual election result is loading.  Your and your boss will be retired. & Eng. & hateful & contempt \\
& The way pitobi failed you will also failed woefully & CDW & hateful & abusive \\
& Bro, go to the field and gather momentum. Social media can only do so much & Eng. & normal & contempt \\
& Thumb  to the working Governor! & Eng. & normal & abusive \\
\bottomrule
\end{tabular}

  \vspace{-1mm}
  \caption{ Examples of correct and incorrect predictions.  \looseness-1}
  \label{tab:error2}
  \end{center}
\end{table*}

\end{document}